\documentclass[10pt,twocolumn,letterpaper]{article}

\usepackage{cvpr}              % To produce the CAMERA-READY version
\definecolor{cvprblue}{rgb}{0.21,0.49,0.74}
\usepackage[pagebackref,breaklinks,colorlinks,allcolors=cvprblue]{hyperref}
\usepackage{tabularray}
\usepackage{graphicx}
\usepackage{float}
\usepackage{algorithm}
\usepackage{algorithmic}
\usepackage{tikz}
\usetikzlibrary{shapes.geometric, arrows.meta, positioning, fit, backgrounds}
\usepackage{multirow}
\usepackage{graphicx}
\usepackage[table,xcdraw]{xcolor}

\def\paperID{****} % *** Enter the Paper ID here
\def\confName{CVPR}
\def\confYear{2026}

\title{NTIRE 2026 Challenge on Bitstream-Corrupted Video Restoration:\\Methods and Results}

\author{Wenbin Zou\thanks{W. Zou, T. Liu, K. Wu, H. Zhuang, Z. Wu, Z. Zhou, R. Timofte, K. Yap, L. Chau and Y. Wang are the NTIRE 2026 chalenge organizers, while the other authors are participants in this challenge. Each team described their own method in the report. Appendix A contains the authors’ teams and affiliations. NTIRE 2026 webpage: \href{https://cvlai.net/ntire/2026}{https://cvlai.net/ntire/2026}.} \quad 
Tianyi Liu\textsuperscript{*} \quad Kejun Wu\textsuperscript{*} \quad Huiping Zhuang\textsuperscript{*} \quad Zongwei Wu\textsuperscript{*} \quad Zhuyun Zhou\textsuperscript{*} \quad
\\ Radu Timofte\textsuperscript{*} \quad Kim-Hui Yap\textsuperscript{*} \quad Lap-Pui Chau\textsuperscript{*} \quad Yi Wang\textsuperscript{*}\thanks{Corresponding author: Yi Wang. \href{mailto:yi-eie.wang@polyu.edu.hk}{yi-eie.wang@polyu.edu.hk}} \quad Shiqi Zhou \quad Xiaodi Shi \quad
\\ Yuxiang Chen \quad Yilian Zhong \quad Shibo Yin \quad Yushun Fang \quad Xilei Zhu \quad Yahui Wang \quad 
\\ Chen Lu \quad Zhitao Wang \quad Lifa Ha \quad Hengyu Man \quad Xiaopeng Fan \quad Priyansh Singh \quad
\\ Sidharth \quad Krrish Dev \quad Soham Kakkar \quad Vinit Jakhetiya \quad Ovais Iqbal Shah \quad Wei Zhou \quad
\\ Linfeng Li \quad Qi Xu \quad Zhenyang Liu \quad Kepeng Xu \quad Tong Qiao \quad Jiachen Tu \quad Guoyi Xu \quad
\\ Yaoxin Jiang \quad Jiajia Liu \quad Yaokun Shi
}

\begin{document}
% \begin{figure*}[t]
%     \centering\includegraphics[width=17cm, height=8.5cm]{example-image-a}
%     \centering\caption{\textbf{(a).} Illustration of the acceleration dimensions in Diffusion Transformers (DiTs). Unlike existing methods, the proposed Heterogeneous Caching (HetCache) jointly models denoising-step redundancy in the diffusion process and token redundancy within the Transformer architecture. \textbf{(b).} By introducing a heterogeneous strategy tailored for video editing, HetCache accelerates diffusion-based video editing while maintaining generation quality.}
%     \label{fig:sptiotemporal similarity}
% \end{figure*}
\maketitle

% \twocolumn[{%
% \renewcommand\twocolumn[1][]{#1}%
% \maketitle
% \begin{center}
%     \centering
%     \captionsetup{type=figure}
%     \includegraphics[width=17cm, height=6cm]{example-image-a}
%     \captionof{figure}{\textbf{(a).} Illustration of the acceleration dimensions in Diffusion Transformers (DiTs). Unlike existing methods, the proposed Heterogeneous Caching (HetCache) jointly models denoising-step redundancy in the diffusion process and token redundancy within the Transformer architecture. \textbf{(b).} By introducing a heterogeneous strategy tailored for video editing, HetCache accelerates diffusion-based video editing while maintaining generation quality.}
% \end{center}%
% }]

\begin{abstract}
This paper reports on the NTIRE 2026 Challenge on Bitstream-Corrupted Video Restoration (BSCVR). The challenge aims to advance research on recovering visually coherent videos from corrupted bitstreams, whose decoding often produces severe spatial-temporal artifacts and content distortion. Built upon recent progress in bitstream-corrupted video recovery, the challenge provides a common benchmark for evaluating restoration methods under realistic corruption settings. We describe the dataset, evaluation protocol, and participating methods, and summarize the final results and main technical trends. The challenge highlights the difficulty of this emerging task and provides useful insights for future research on robust video restoration under practical bitstream corruption.
\end{abstract}    
\section{Introduction}

Video bitstreams are vulnerable to corruption during transmission, storage, and decoding in real-world multimedia systems. Even minor packet loss, bit errors, or damaged bitstream segments may lead to severe spatial-temporal degradation after decoding, causing mixed visual artifacts, content distortion, and temporal inconsistency. This corruption not only degrades the user experience, but also affects the reliability of downstream video applications in surveillance, streaming, communication, and visual analytics.

Compared with conventional video restoration problems such as denoising, deblurring, and compression artifact reduction, bitstream-corrupted video recovery is more challenging because the resulting degradation is often irregular, non-stationary, and highly dependent on codec behavior and inter-frame prediction. Traditional restoration methods usually assume relatively stable degradation priors, while video inpainting and error concealment methods often rely on manually designed masks or simplified missing-pattern assumptions. These settings are insufficient to faithfully reflect realistic bitstream corruption, where corrupted regions may contain mixed residual information, complex artifact patterns, and propagation across frames.

Recent studies have started to establish bitstream-corrupted video recovery as a dedicated research problem. Liu \etal introduced the first large-scale benchmark dataset, BSCV, together with a prototypical recovery baseline, demonstrating that realistic corruption decoded from corrupted bitstreams differs substantially from conventional manually simulated masks and poses new challenges to existing video recovery methods \cite{liu2024bitstream}. Later, Wang \etal~\cite{wang2025blind} and Liu \etal~\cite{liu2025towards} further explored enhanced recovery and explored blind setting. These works provide an important foundation for the community, but the problem remains highly challenging in terms of corruption diversity, recovery quality, and practical deployment.

\begin{figure*}[ht]
    \centering
    \includegraphics[width=1\linewidth]{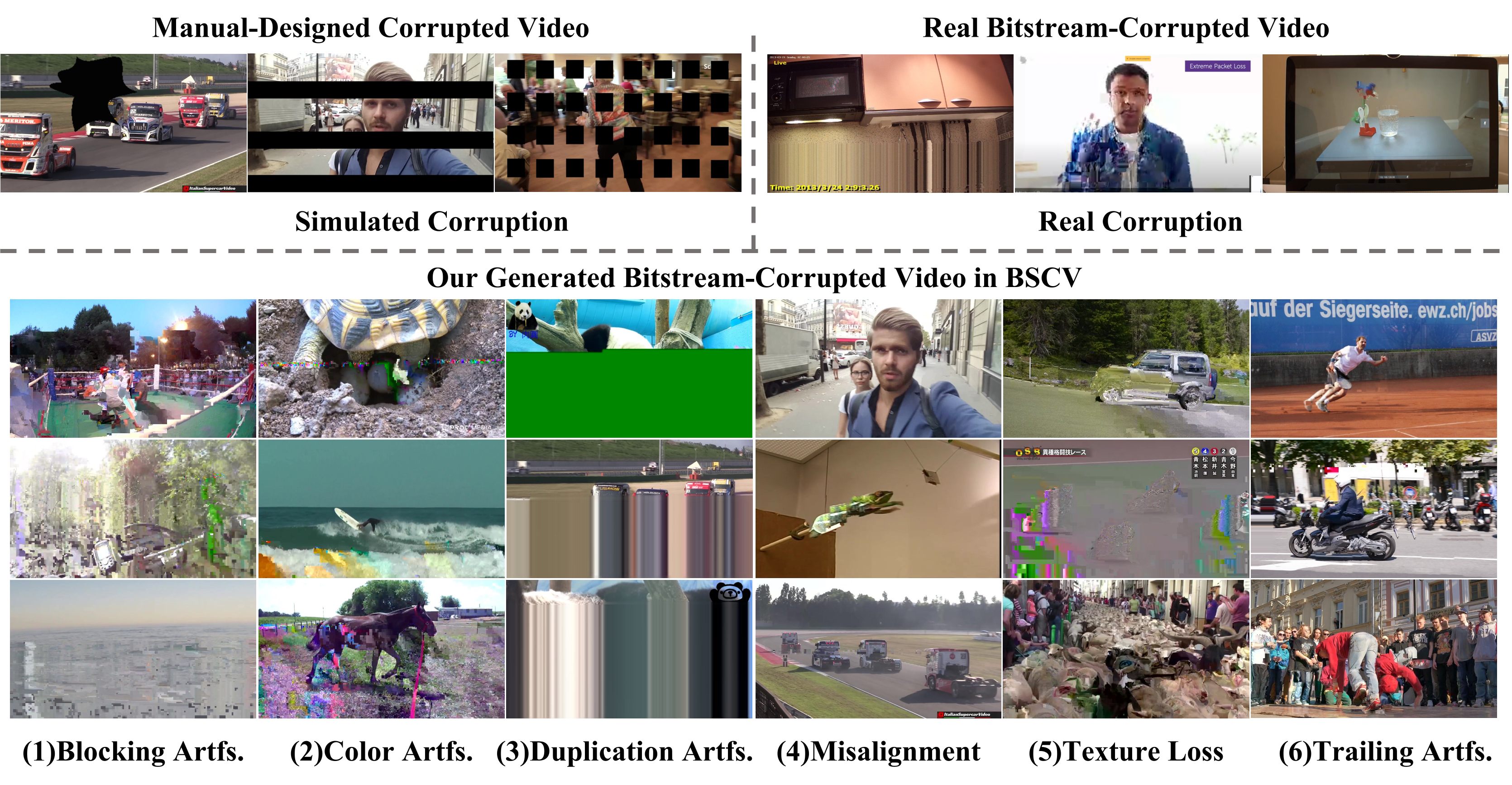}
    \caption{Video corruption pattern in bitstream-corrupted video recovery problem summarized by~\cite{liu2024bitstream}. Compared with the simulated video corruption in existing inpainting or error concealment research, BSCVR contains various realistic corruption patterns including (1) block artifacts (artfs.), (2) color artifacts, (3) duplication artifacts, (4) misalignment, (5) texture loss, (6) trailing artifacts, which is closer to the corrupted video in real world.}
    \label{fig:dataset}
\end{figure*}

To encourage further progress in this emerging area, we organize the NTIRE 2026 Challenge on Bitstream-Corrupted Video Restoration (BSCVR). This challenge aims to provide a common testbed for evaluating restoration methods under realistic bitstream corruption, and to benchmark current solutions in terms of reconstruction fidelity, perceptual quality, and robustness under diverse corruption patterns. By bringing together researchers working on video restoration, inpainting, generative enhancement, and codec-aware recovery, the challenge offers a timely opportunity to advance state-of-the-art methods and to better understand the evolving technical trends for bitstream-corrupted video restoration.

This challenge is one of the challenges associated with the NTIRE 2026 Workshop~\footnote{\url{https://www.cvlai.net/ntire/2026/}} on:
deepfake detection~\cite{ntire26deepfake}, 
high-resolution depth~\cite{ntire26hrdepth},
multi-exposure image fusion~\cite{ntire26raim_fusion}, 
AI flash portrait~\cite{ntire26raim_portrait}, 
professional image quality assessment~\cite{ntire26raim_piqa},
light field super-resolution~\cite{ntire26lightsr},
3D content super-resolution~\cite{ntire263dsr},
bitstream-corrupted video restoration,
X-AIGC quality assessment~\cite{ntire26XAIGCqa},
shadow removal~\cite{ntire26shadow},
ambient lighting normalization~\cite{ntire26lightnorm},
controllable Bokeh rendering~\cite{ntire26bokeh},
rip current detection and segmentation~\cite{ntire26ripdetseg},
low light image enhancement~\cite{ntire26llie},
high FPS video frame interpolation~\cite{ntire26highfps},
Night-time dehazing~\cite{ntire26nthaze,ntire26nthaze_rep},
learned ISP with unpaired data~\cite{ntire26isp},
short-form UGC video restoration~\cite{ntire26ugcvideo},
raindrop removal for dual-focused images~\cite{ntire26dual_focus},
image super-resolution (x4)~\cite{ntire26srx4},
photography retouching transfer~\cite{ntire26retouching},
mobile real-word super-resolution~\cite{ntire26rwsr},
remote sensing infrared super-resolution~\cite{ntire26rsirsr},
AI-Generated image detection~\cite{ntire26aigendet},
cross-domain few-shot object detection~\cite{ntire26cdfsod},
financial receipt restoration and reasoning~\cite{ntire26finrec},
real-world face restoration~\cite{ntire26faceres},
reflection removal~\cite{ntire26reflection},
anomaly detection of face enhancement~\cite{ntire26anomalydet},
video saliency prediction~\cite{ntire26videosal},
efficient super-resolution~\cite{ntire26effsr},
3d restoration and reconstruction in adverse conditions~\cite{ntire26realx3d},
image denoising~\cite{ntire26denoising},
blind computational aberration correction~\cite{ntire26aberration},
event-based image deblurring~\cite{ntire26eventblurr},
efficient burst HDR and restoration~\cite{ntire26bursthdr},
low-light enhancement: `twilight cowboy'~\cite{ntire26twilight},
and efficient low light image enhancement~\cite{ntire26effllie}. In this report, we present the challenge setup, datasets and evaluation protocol, summarize the participating methods, and analyze the final results and main technical trends observed from the submitted solutions.

% \usepackage{multirow}
% \usepackage{graphicx}
\begin{table*}[ht]
\centering
\caption{NTIRE 2026 Bitstream-Corrupted Video Restoration Challenge results, final rankings, and the main characteristics of the solutions. Note that, the average PSNR value achieved on the test set is used for final ranking.}
\label{tab:my-table}
\resizebox{\textwidth}{!}{%
\begin{tabular}{c|ccccccc}
\hline
                                    & \multicolumn{7}{c}{NTIRE 2026 Challenge on Bitstream-Corrupted Video Restoration Results}                   \\ \cline{2-8} 
\multirow{-2}{*}{Team}              & Codabench User                     & PSNR$\uparrow$ (Primary) & SSIM$\uparrow$   & LPIPS$\downarrow$   & Rank PSNR & Rank SSIM & Final Rank \\ \hline
{\color[HTML]{13181D} MGTV-AI}      & {\color[HTML]{13181D} nerror}      & 33.642        & 0.9334 & 0.0900 & 1         & 2         & 1          \\
{\color[HTML]{13181D} RedMediaTech} & {\color[HTML]{13181D} chenyuxiang} & 32.865         & 0.9344 & 0.0852  & 2         & 1         & 2          \\
{\color[HTML]{13181D} bighit}       & {\color[HTML]{13181D} hyena}       & 27.873        & 0.8933 & 0.1388  & 3         & 3         & 3          \\
{\color[HTML]{13181D} Vroom}        & {\color[HTML]{13181D} priyansh}    & 27.370        & 0.8713 & 0.2028  & 4         & 5         & 4          \\
{\color[HTML]{13181D} weichow}      & {\color[HTML]{13181D} weichow}     & 27.276        & 0.8727 & 0.1866  & 5         & 4         & 5          \\
{\color[HTML]{13181D} holding}      & {\color[HTML]{13181D} zyliu}       & 26.889        & 0.8724 & 0.1657  & 6         & 6         & 6          \\
{\color[HTML]{13181D} NTR}          & {\color[HTML]{13181D} miketjc}     & 25.840        & 0.8262 & 0.2741  & 7         & 7         & 7          \\ \hline
\end{tabular}%
}
\label{tab:results}
\end{table*}

\section{Related Works}

\subsection{Traditional Methods}

\noindent \textbf{Video restoration} has been extensively studied for degradations such as blur and low resolution~\cite{nah2019ntire}, noise~\cite{zhang2021plug}, compression artifacts~\cite{kwon2015efficient}, where the underlying corruption is usually modeled with relatively stable priors. 
Representative methods exploit temporal alignment~\cite{tian2020tdan, chan2022basicvsr++}, feature fusion~\cite{wang2019edvr, yi2019progressive}, and transformer-based modeling~\cite{liang2022rvrt, liang2024vrt} to recover clean video content from degraded observations. 
However, these methods are generally designed for global or statistically regular degradation and are not well suited to bitstream corruption, whose decoded artifacts often have irregular temporal propagation.

\noindent \textbf{Video error concealment} is a classical post-decoding solution for repairing corrupted regions in decoded frames. 
Earlier methods estimate missing contents using spatial interpolation~\cite{wang1998error, joint2003draft}, temporal motion compensation~\cite{zhang2024swin}, or hybrid strategies~\cite{ye2008hybrid}. 
More recent learning-based approaches also improve recovery quality under simulated stripe-like or block-like corruption patterns~\cite{8682097}. 
Nevertheless, most of them still rely on handcrafted assumptions about missing regions and do not explicitly address the complex corruption patterns caused by realistic bitstream damage.

\noindent \textbf{Video inpainting} is closely related because it aims to fill missing regions using spatial-temporal context from neighboring frames. 
Modern methods, especially flow-guided and transformer-based approaches, have substantially improved the quality of video completion under arbitrary masks~\cite{xu2019deep, li2022towards, zhou2023propainter}. 
As a result, video inpainting has become an important baseline for corrupted video recovery. 
However, existing inpainting methods are usually developed with simulated masks and often assume fully missing regions, whereas bitstream-corrupted videos typically contain partially preserved but misleading residual content, irregular artifact shapes, and more complicated temporal propagation. 
This gap limits the effectiveness of directly applying existing inpainting frameworks to bitstream-corrupted video restoration.

\subsection{Bitstream-Corrupted Video Recovery}
Bitstream-corrupted video recovery has only recently emerged as a dedicated research topic. Liu \etal first introduced BSCV, the first large-scale benchmark for this task, and proposed a prototypical recovery framework that leverages residual visual cues within corrupted regions together with neighboring spatial-temporal context for recovery \cite{liu2024bitstream}.
This work established the task setting and showed that realistic corruption decoded from corrupted bitstreams differs fundamentally from conventional mask-based simulation used in error concealment and video inpainting.

Building on this line of research, Liu \etal further investigated improved bitstream-corrupted video recovery guided by visual foundation models~\cite{liu2025towards}. 
Wang \etal proposed the integration of diffusion priors to achieve a more robust restoration~\cite{wang2025blind}.
Their method integrates external priors and knowledge into a recovery framework to perform corruption localization and completion of a corruption-aware feature, demonstrating the feasibility of moving toward more practical and deployable recovery systems.

Despite these advances, bitstream-corrupted video restoration remains far from solved. 
Existing benchmarks and methods still face challenges in handling diverse corruption patterns, achieving robust recovery under practical conditions, and generalizing across different corrupted contents and artifact forms. 
Therefore, a standard challenge benchmark is valuable for systematically evaluating current methods, comparing technical designs, and identifying promising directions for future research.
\section{NTIRE 2026 Challenge}
In this section, we introduce the NTIRE 2026 Bitstream-Corrupted Video Restoration Challenge. We first introduce the official datasets and toolbox of this challenge. Then, we review two phases of this challenge. Finally, we summarize the common trends in the submitted solutions.

\begin{figure*}
    \centering
    \includegraphics[width=1\linewidth]{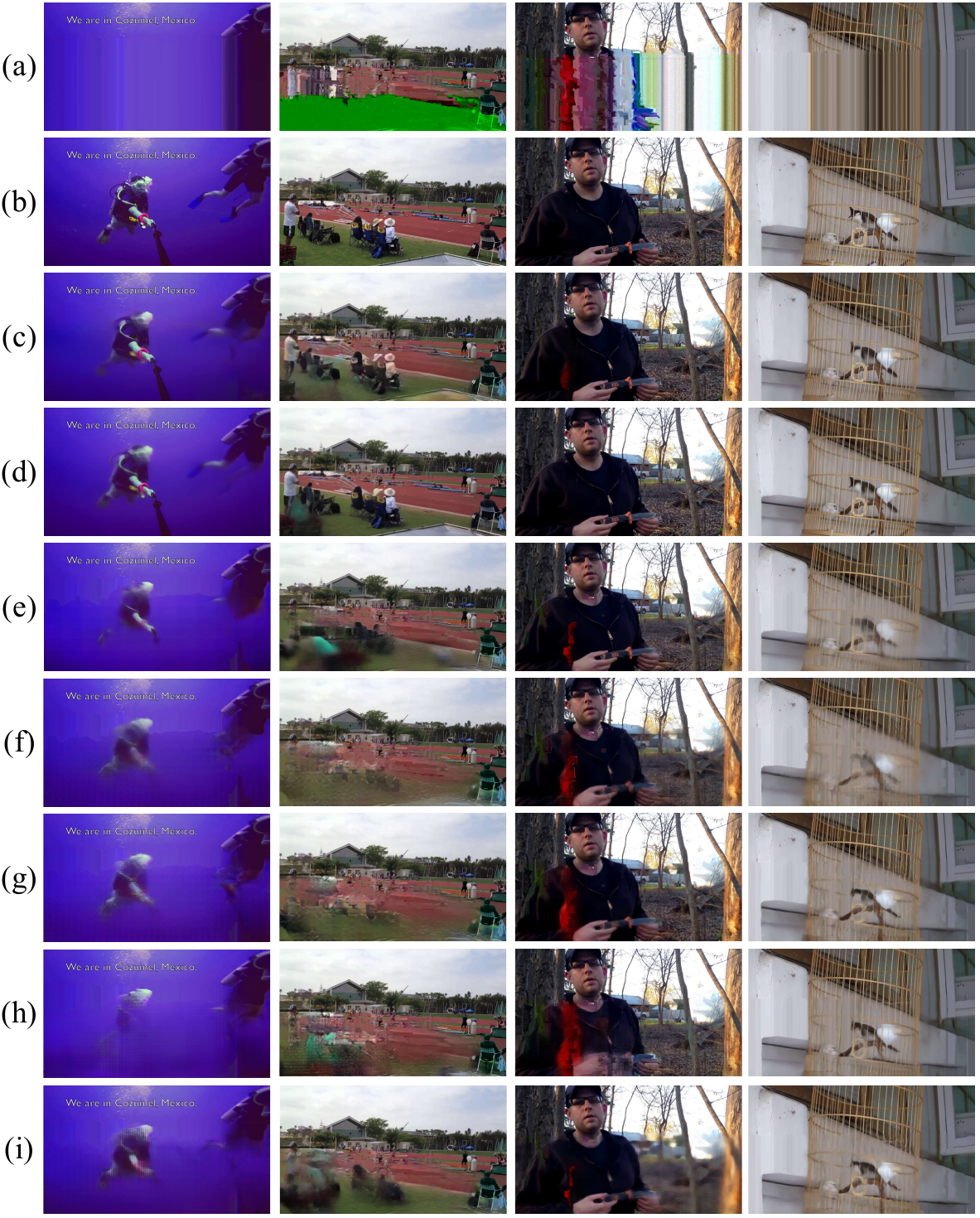}
    \caption{Visualization of the final results of participants, along with corresponding input reference frames and GT images. Each column represents a video of a different scene. Among them, (a-i) respectively represent the results of Bitstream Corrupted Input, GT, MGTV-AI team, RedMediaTech team, Bighit team, Vroom team, Weichow team, Holding team, and NTR team.}
    \label{fig:results}
\end{figure*}

\subsection{Datasets, Toolbox and Evaluation}

\textbf{Training set.} The training set comprises 3,471 bitstream-corrupted videos in HD resolution, acquired via the simulation pipeline from the BSCV benchmark \cite{liu2024bitstream}, as shown in Figure \ref{fig:dataset}. These videos mainly depict general scenes and contain decoding artifacts with non-uniform spatio-temporal distributions. For this set, we release the bitstream-corrupted (BSC) frame sequences, the corresponding uncorrupted ground-truth (GT) sequences, and per-frame binary mask sequences. The masks indicate corrupted regions and are computed using difference maps between corrupted and uncorrupted videos, followed by threshold suppression and morphological filtering to simulate human indication. Challenge participants are permitted to use additional training data and pretrained networks, provided that detailed sources and amounts are specified in the final submission. Furthermore, there are no restrictions on model size or running time, though these metrics must be explicitly reported. 

\noindent\textbf{Validation and test set.} The validation set consists of 50 video clips, for which all corresponding data (BSC, GT, and masks) are fully released. For the hidden test set used for final ranking, only the corrupted videos and their corresponding mask sequences are provided. 

\noindent\textbf{Toolbox.} We provide a development toolkit to facilitate beginner competitors in quickly getting access to the challenge. It serves as a baseline, supporting the minimum re-implementation and re-training of B2SCVR \cite{liu2025towards} using a single 24GB VRAM 3090 GPU with a batch size of 1. Details regarding the development environment can be found in the associated B2SCVR GitHub repository: \href{https://github.com/LIUTIGHE/B2SCVR}{https://github.com/LIUTIGHE/B2SCVR}.

\noindent\textbf{Evaluation.} We use the standard Peak Signal To Noise Ratio (PSNR) and, complementary, the Structural Similarity (SSIM) index as often employed in the literature. PSNR and SSIM implementations are found in most of the image processing toolboxes. We first calculate the average results over bitstream-corrupted frames in a video, and then average the results among all videos in the validation/test dataset. In this challenge, the final result is ranked by normalizing PSNR and SSIM in the RGB domain and then weighting them.

\subsection{Challenge Phases}
\textbf{Development Phase.} The participants can download the validation set and apply their developed models to the bitstream-corrupted and ground turth video pairs to generate their clear versions. A validation leaderboard is available during this phase. The participants can compare their scores with the ones achieved by the baseline models or models developed by other participants.

\noindent\textbf{Test phase.} The participants are required to apply their models to the released test set, and submit their clear output video sequences to the test server. The test server is available online during this phase, and will be closed after the test deadline. The participants are asked to submit the clear output video sequences results, codes, and a fact sheet of their methods before the given deadline.

\subsection{Challenge Results}
Among the 153 registered participants, 7 teams have participated in the final test phase of the NTIRE 2026 Bitstream-Corrupted Video Restoration Challenge and submitted their results, codes, and factsheets. 

Table \ref{tab:results} reports the PSNR, SSIM, and LPIPS scores achieved by these methods. Notably, team MGTV-AI achieved the highest quantitative restoration performance, securing the best overall PSNR of 33.6423 dB and an overall SSIM of 0.9334. Meanwhile, team RedMediaTech excelled in perceptual quality, obtaining the best overall LPIPS score of 0.0852. By utilizing the Wan2.1 base network, team redmediatech demonstrated the effectiveness of strong generative priors in producing perceptually pleasing reconstructions for severely corrupted videos.

In addition, the Figure \ref{fig:results} presents a comparative analysis of video restoration performance across various competitive teams, evaluating their ability to reconstruct frames from severely degraded bitstream-corrupted inputs. The evaluation is structured across nine rows (a–i), representing the input, the ground truth, and seven distinct algorithmic approaches applied to four different video scenarios. Among the participants, the MGTV-AI (c) and RedMediaTech (d) teams demonstrate superior restoration capabilities, effectively neutralizing bitstream errors while maintaining sharp edges and high-frequency details. The mid-tier results from the Bighit (e),  Vroom (f), weichow (g), holding(h), and NTR (i) teams show successful artifact suppression, but they struggle with "softness" or slight blurring in the output. While the heavy blockiness is removed, these methods often fail to perfectly reconstruct fine textures, such as the facial features of the speaker or the intricate wires of the birdcage. Overall, the comparison highlights that while modern AI-driven restoration can effectively "hallucinate" missing data to repair structural damage, the primary challenge remains the accurate recovery of semantic details (like text) and the maintenance of temporal stability across frames where the original bitstream data is almost entirely lost.

Across all the submissions, a significant observation is the widespread adoption of the B2SCVR architecture~\cite{liu2025towards} as a robust baseline, alongside the integration of visual foundation models. Among the 6 submitted solutions, 3 teams (weichow, Vroom, and holding) built their frameworks upon the B2SCVR base network. Another prominent trend is the extensive use of external semantic and structural priors (such as SAM2~\cite{ravi2024sam2}, DINO~\cite{dinov2, dinov3}, and Qwen-Image VAE~\cite{qwen}) coupled with Parameter-Efficient Fine-Tuning (PEFT) techniques~\cite{peft}. Specifically, half of the teams utilized PEFT methods, such as LoRA~\cite{lora} or MoE-LoRA~\cite{mola}, to adapt these large models efficiently. This trend highlights the generalizability of foundation models and their effectiveness in enhancing video restoration performance while managing computational complexity and parameter counts.

We briefly describe these solutions in Section 4, and introduce the corresponding team members in Appendix 5.

\begin{figure*}[t]
    \centering
    \includegraphics[width=\textwidth]{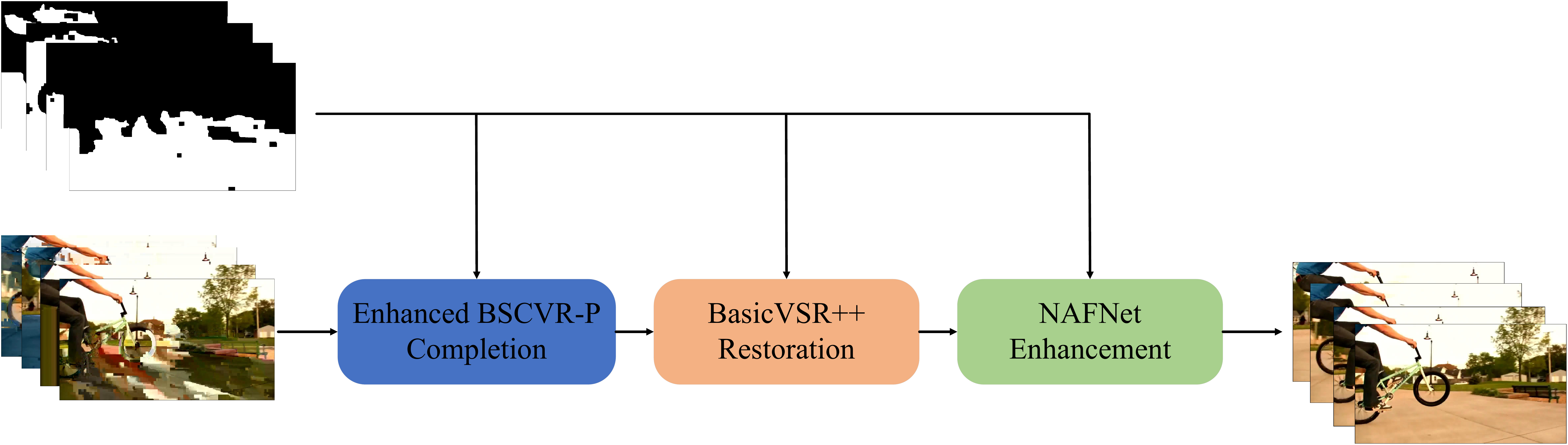} 
    \caption{The MGTV-AI Team: Three-Stage Framework For BitStream-corrupted Video Restoration.}
    \label{fig:mgtv}
\end{figure*}

\section{Challenge Teams and Methods}

\subsection{MGTV-AI}
\textbf{General method description.}
The MGTV-AI team has proposed a three-stage framework for the Bitstream Corrupted Video Restoration task, as shown in Figure~\ref{fig:mgtv}. The motivation for this method is that the official provided video mask does not fully cover all damaged areas, resulting in the inability to repair the damaged parts outside the mask, and a single completion network is difficult to achieve satisfactory results. In the first stage, the team focuses on completing local and rough information within the masked area. To achieve this goal, they used an optimized version of BSCVR-P for completing mask regions. Specifically, they replaced the optical flow prediction network in BSCVR-P with ProPainter's network structure~\cite{liu2024bitstream, propainter} and retrained it on the provided training data. In addition, they also extended the feature enhancer module and the temporal focal transformer module to further improve performance. The second stage focuses on global and temporal video restoration. The team takes the output results of the first stage and the mask as input, applies BasicVSR++ \cite{chan2022basicvsr++} video repair network, and refines and repairs the damaged areas inside and outside the mask to achieve better time alignment effect. The third stage focuses on enhancing global and spatial information. Based on the output results and masks of the second stage, the team used the image restoration network NAFNet~\cite{nafnet} to refine the output of the previous stage, in order to enhance and restore more image details.

\textbf{Implementation and Training details.}
In the first stage, the model adopts L1 and T-PatchGAN~\cite{chang2019free} loss functions, and uses Adam optimizer for a total of 700k iterations of training, with a batch size of 4 and an initial learning rate of 1e-4. After 400k iterations, the weight of L1 loss is increased to 10. During the training period, the video was adjusted to a resolution of $640 \times360 $, supplemented by random horizontal flipping for data augmentation, and PyTorch's FSDP technology was used to reduce GPU memory consumption. In terms of data sampling, the number of local frames and reference frames are set to 5 and 10 respectively, and a sliding window strategy is applied to ensure data diversity, where local frames are only sampled from frames with mask ratios greater than zero. When inferring, the input image will first be adjusted to $640 \times360 $, then the result will be adjusted back to the original resolution, and finally fused based on the mask. 

In the second stage, the model uses Adam optimizer and cosine annealing learning rate scheduling. The initial learning rates of the main network and optical flow network are set to $1 \times10-4 $and $2.5 \times10-5 $, respectively, with a batch size of 8. The data sampling strategy is consistent with the local grid sampling method used in the first stage. The training process is divided into two steps: the model is first trained 200k iterations at a resolution of $256 \times256 $, and then fine tuned 50k iterations at a resolution of $512 \times512 $. 

In the third stage, the model also uses the Adam optimizer for a total of 400k iterations of training, with an initial learning rate of 1e-3, and gradually decreases to 1e-6 through cosine annealing scheduling. The patch size during training is $256 \times256 $, and the batch size is 32. In this stage, only frames with mask ratios greater than zero are used, and the training blocks are randomly cropped from the output results of the previous stage. 

In terms of testing and model fusion, the inference process in the second and third stages is carried out at the original resolution. To further improve performance, the team adopted an ensemble strategy similar to BasicVSR++~\cite{chan2022basicvsr++} in the latter two stages, weighting the original input and its horizontally flipped version of the predicted results, and fusing them into the final output. The fusion output of the third stage serves as the final evaluation result.

\begin{figure*}
    \centering
    \includegraphics[width=\linewidth]{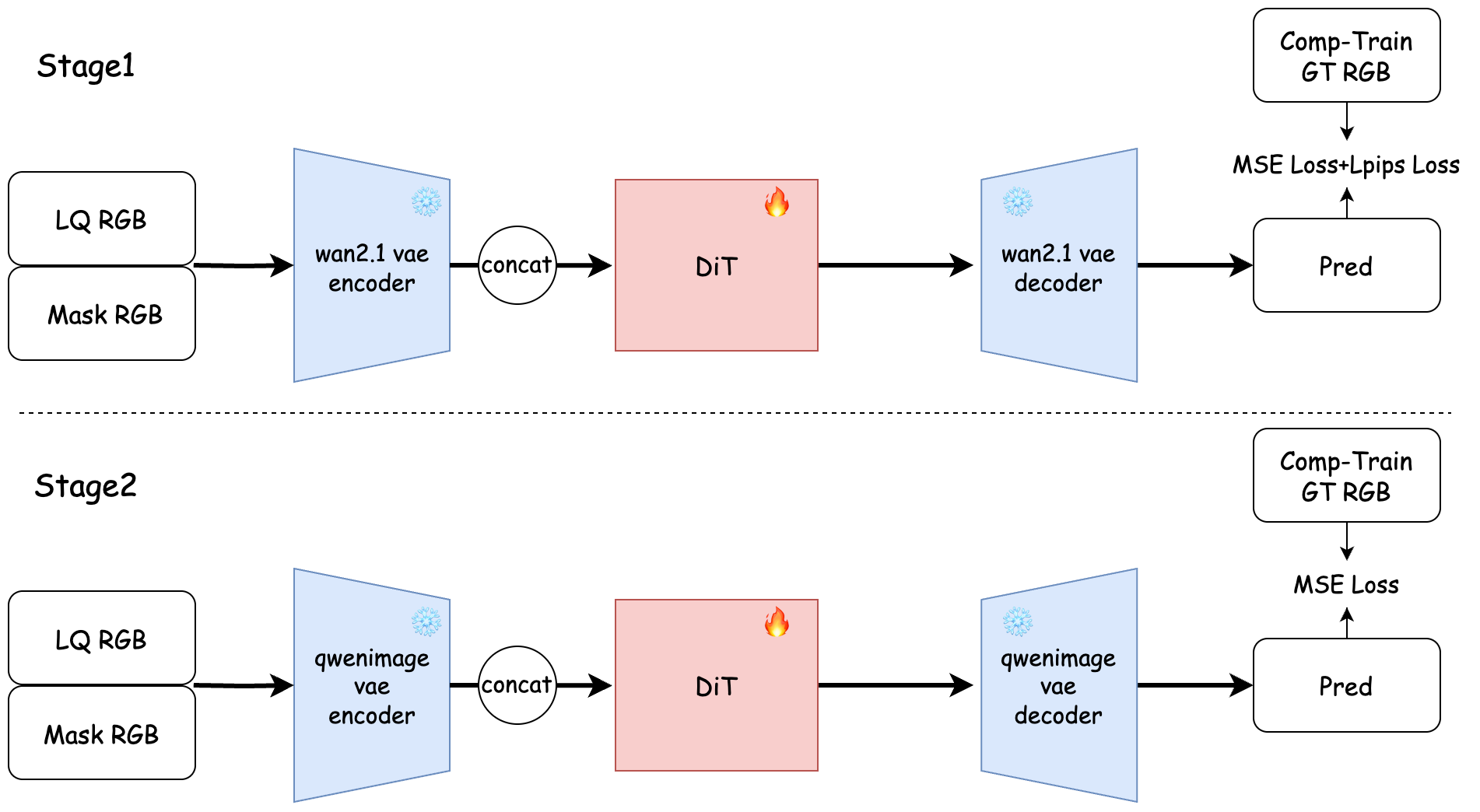}
    \caption{The RedMediaTech team proposed a single-step video restoration framework based on Wan2.1 DiT with a two-stage training strategy. The first stage leverages MSE and LPIPS losses to exploit strong generative priors for perceptual quality, while the second stage fine-tunes with MSE loss to improve distortion metrics (PSNR/SSIM). Additionally, replacing the original VAE with Qwen-Image VAE enhances robustness to large motion and complex temporal variations.}
    \label{fig:redmedia}
\end{figure*}

\subsection{RedMediaTech}
\textbf{General method description.}
The RedMediaTech team proposes a single-step video restoration framework (Figure \ref{fig:redmedia}) built upon the Wan2.1 Diffusion Transformer (DiT) \cite{wan} to address the Bitstream-Corrupted Video Restoration Challenge. To more effectively handle large-motion scenarios and complex temporal variations, the method modifies the base architecture by replacing the original Wan2.1 Variational Autoencoder (VAE) with the Qwen-Image VAE \cite{qwen}, which provides an enhanced representation capacity for corrupted video features.

The core of the proposed approach lies in a tailored two-stage training strategy designed to strike an optimal balance between perceptual quality and distortion-oriented metrics. In the first stage, the network is optimized using a joint loss function comprising Mean Squared Error (MSE) and LPIPS. By leveraging the strong generative priors inherent in the Wan model, this initial phase enables rapid convergence on the restoration task while establishing high perceptual quality. In the second stage, the framework is fine-tuned exclusively with the MSE loss. This targeted refinement shifts the optimization focus to maximize objective distortion-based metrics, such as PSNR and SSIM, ensuring high-fidelity structural reconstruction of the corrupted bitstream videos.

\textbf{Implementation details.}
The team employs a two-stage training strategy to optimize the model. In the first stage, the network is trained using a joint loss function comprising Mean Squared Error (MSE) and LPIPS to balance distortion metrics and perceptual quality. Given a sequence of low-quality RGB frames and their corresponding mask frames, the original VAE encoder of Wan~\cite{wan} is utilized to extract their respective latent representations. These latents are concatenated along the channel dimension and fed into the network to predict a latent residual flow field. The restored latent is obtained by subtracting this predicted residual from the low-quality input latent. Finally, the restored latent is decoded back into the RGB space using the VAE decoder. The overall reconstruction during this stage is supervised by the combined MSE and LPIPS loss.

In the second stage, the training focuses on addressing the limitations of the original Wan2.1 VAE, which experiences a drop in encoding capability when handling consecutive corrupted frames and large-motion scenes due to the disruption of redundant temporal priors. To resolve this, the Wan2.1 VAE is replaced by the Qwen-Image VAE. This substitution enhances the latent representation capacity, allowing the model to better preserve fine details and reduce artifacts in sequences with severe corruption or rapid motion. Because the Qwen-Image VAE does not utilize temporal compression, the Diffusion Transformer (DiT)~\cite{DiT} processes approximately four times as many tokens, requiring higher computational and memory resources. Following this VAE upgrade, the model is fine-tuned exclusively with the MSE loss. By focusing solely on pixel-wise reconstruction errors, this final stage emphasizes distortion-oriented metrics, further improving PSNR and SSIM for high-fidelity, temporally consistent frame recovery.

\begin{figure*}[ht]
\centering
\includegraphics[width=1\linewidth]{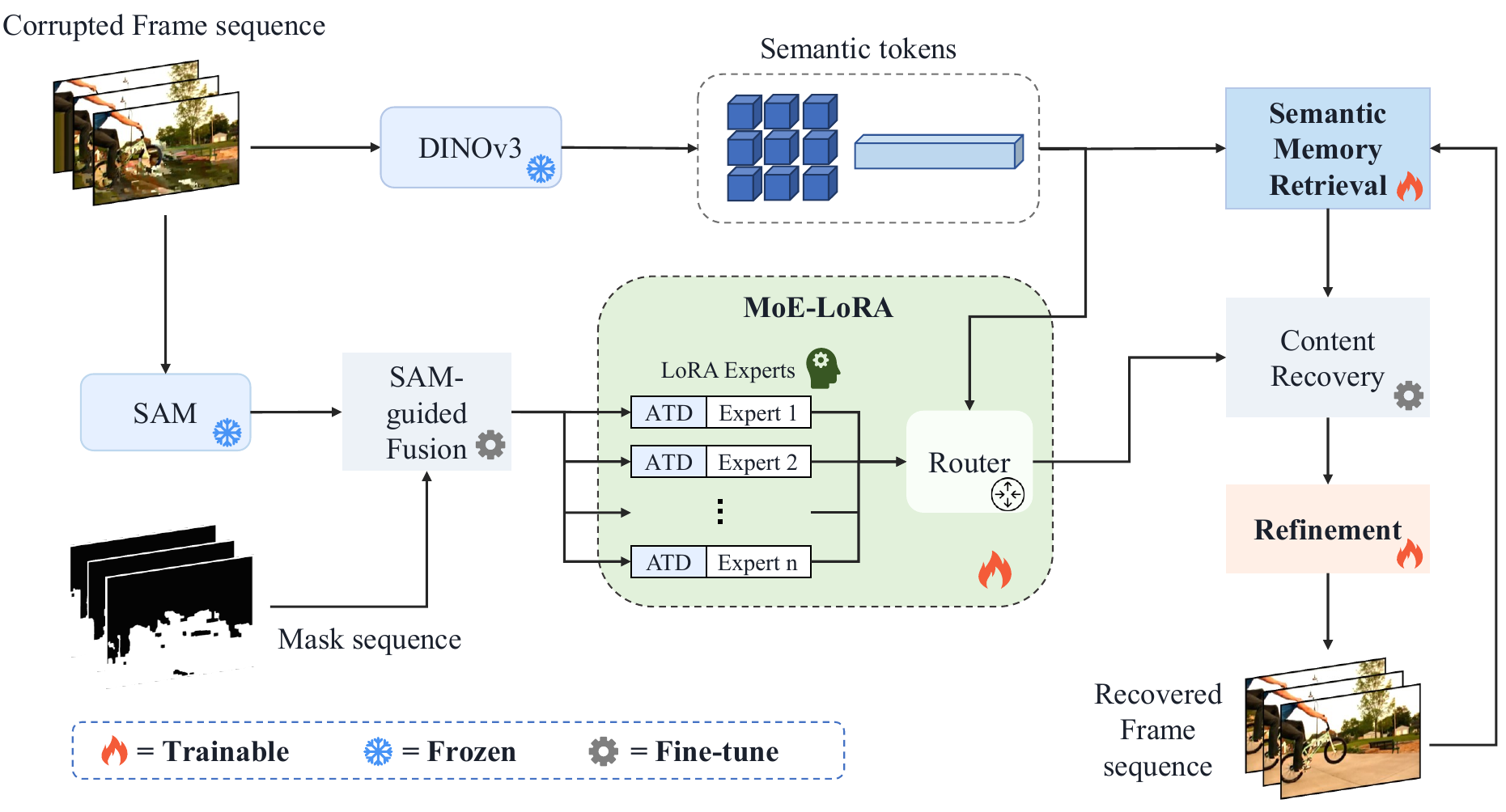}
\caption{
The bighit team proposed two-stage framework for bitstream-corrupted video restoration. Corrupted frame sequences and masks are first fused with SAM2 structural embeddings and DINOv3 semantic tokens. A semantic-memory retrieval branch and an router-guided MoE-LoRA adaptation branch are jointly used for stage-1 content recovery. An optional stage-2 refinement module further suppresses residual artifacts and improves boundary consistency to produce the final recovered frame sequence.
}
\label{fig:bighit}
\end{figure*}

\subsection{bighit}
\textbf{General method description.}
The bighit team proposes a two-stage framework for bitstream-corrupted video restoration, featuring a semantic memory bank and a router-guided mixture of LoRA experts (termed MoE-LoRA)~\cite{lora, mola} as its core components. The overall structure of the proposed framework is illustrated in Fig.~\ref{fig:bighit}.

The first stage of the framework is built upon two key architectural designs. First, a semantic memory bank is utilized to retrieve high-level semantic context from DINOv3~\cite{dinov3} features across the video sequence. This mechanism enables the model to recall structurally relevant information from reliable historical frames, thereby enhancing restoration stability beyond the confines of a local temporal window—a feature that is particularly beneficial when the current frame is severely corrupted. Second, the team introduces an MoE-LoRA module, where multiple lightweight LoRA adapters function as dynamic "experts." These experts are adaptively weighted and fused in the feature space based on semantic and degradation-aware cues. This design allows a single restoration backbone to handle heterogeneous corruption patterns effectively, improving parameter efficiency compared to maintaining separate, full-scale models.

These core components are seamlessly integrated into a comprehensive restoration pipeline that includes a convolutional encoder-decoder backbone, SAM2-guided~\cite{ravi2024sam2} structural fusion, bidirectional flow propagation~, and temporal transformer aggregation to ensure robust spatio-temporal recovery. Furthermore, an optional second stage applies a lightweight NAFNet-style enhancer~\cite{nafnet} to further suppress residual artifacts and refine boundary inconsistencies. Overall, the proposed method is engineered to improve structural fidelity and temporal consistency under severe bitstream corruption while maintaining a practical, single-pipeline deployment.

\begin{figure*}[t]
\centering
\includegraphics[width=1\textwidth]{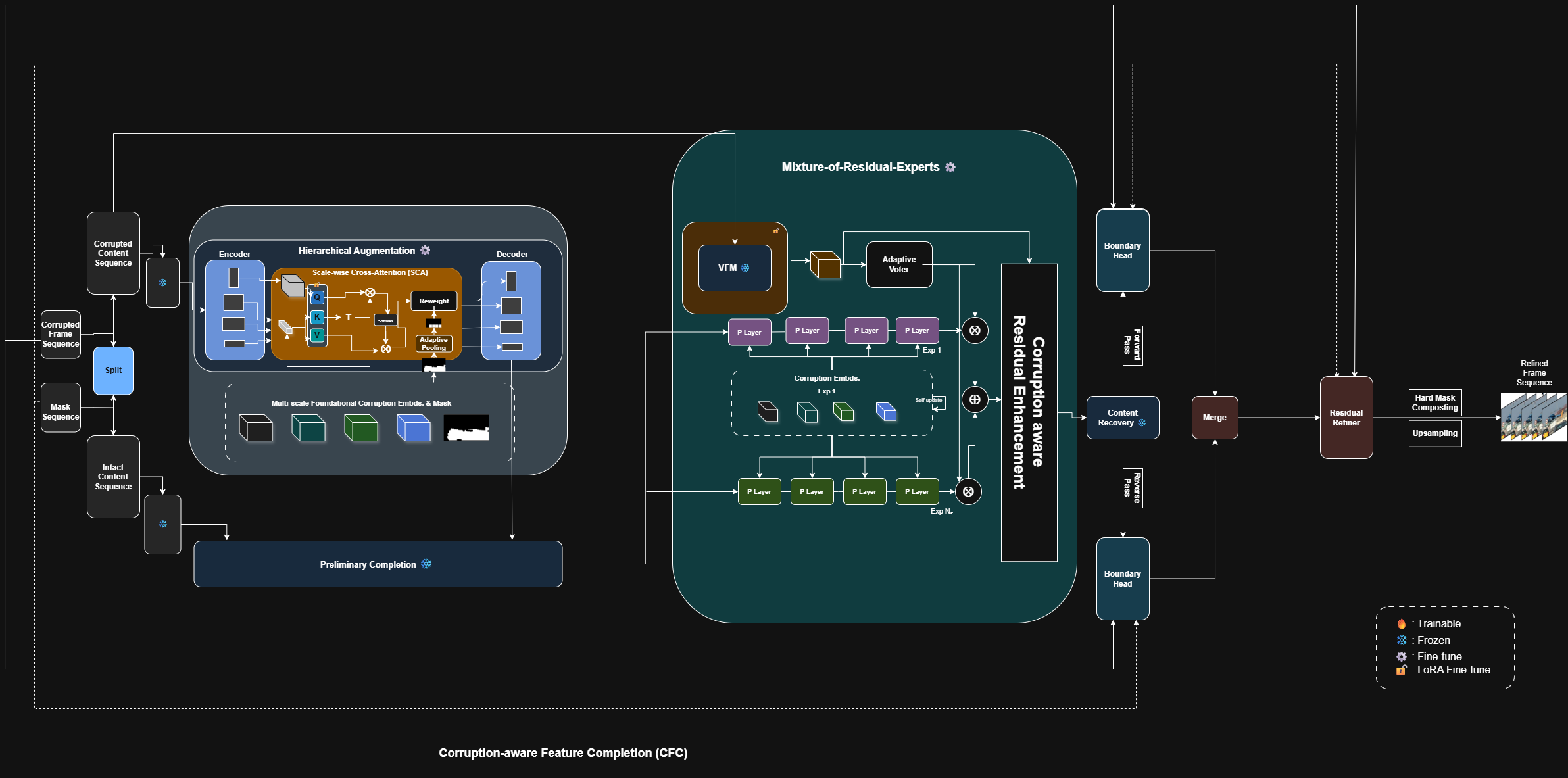}
\caption{The Vroom team: Enhanced B2SCVR: SAM2-Prior Guided Bitstream-Corrupted Video Restoration with LoRA and Boundary Refinement. }
\label{fig:architecture}
\end{figure*}

\textbf{Implementation details.}
The proposed framework utilizes several pre-trained models as robust visual priors, including SAM2 for structural feature extraction and DINOv3 ViT-S/16 for semantic adaptation and memory retrieval~\cite{ravi2024sam2, dinov3}, alongside SPyNet~\cite{spynet} for optical-flow-based temporal propagation. The primary restoration network features an adaptation module equipped with a 4-expert MoE-LoRA routing strategy (with a LoRA rank of 8)~\cite{lora, mola} and a semantic memory bank configured with a capacity of 32, 10 semantic clusters, and top-4 retrieval rules. The optional second-stage enhancer adopts a NAFNet-style residual U-Net architecture~\cite{nafnet, unet} with a base width of 32, an encoder block configuration of $[2, 2, 4, 8]$, a middle depth of 12, and a decoder block configuration of $[2, 2, 2, 2]$. No additional external datasets are used for task-specific training, relying exclusively on the official NTIRE training data.

The training process is conducted in multiple stages. The main restoration network is trained on cropped patches of size $432 \times 240$, utilizing sequences of 5 local frames and 3 reference frames. The model is optimized using the Adam optimizer with an initial learning rate of $1 \times 10^{-4}$, a batch size of 1, and gradient accumulation over 4 steps. The loss function comprises weighted valid and hole reconstruction losses, combined with a perceptual loss (weight set to 0.5); adversarial losses are intentionally disabled. The optimization process spans 100,000 iterations using a cosine-annealing-restart learning rate schedule, followed by a late-stage MSE-oriented refinement phase. For the second-stage enhancer, the initial learning rate is set to $1 \times 10^{-3}$, employing a two-step warm-up and fine-tuning strategy where the enhancer is first trained on corrupted inputs and subsequently fine-tuned on the outputs generated by the stage-1 model.During testing, the default inference pipeline employs a sliding-window approach with per-video memory-bank resetting, optional reference-quality selection, and dilated-mask composition. To optimize computational efficiency and deployment, the framework supports mixed-precision CUDA inference, chunked execution, and low-resolution enhancer inference to reduce the memory footprint, along with optional multi-pass and test-time augmentation (TTA) variants.

\subsection{Vroom}
\textbf{General method description.}
The Vroom team proposes an enhanced video restoration framework built upon the B2SCVR baseline~\cite{liu2025towards}, incorporating several key architectural modifications to improve spatial-temporal recovery and reduce visual artifacts.

To leverage robust semantic and spatial priors, the method integrates a pre-trained SAM2 encoder~\cite{ravi2024sam2} into the restoration pipeline. To maintain parameter efficiency, the SAM2 backbone remains frozen while Low-Rank Adaptation (LoRA) modules~\cite{lora} are injected into its attention layers. The extracted SAM2 feature maps are then integrated into the main B2SCVR restoration backbone via a dedicated, trainable fusion module (SAMFuser). In a subsequent stage, the framework further adapts to the restoration task by injecting LoRA modules into the Temporal Focal Transformer~\cite{li2022towards} of the backbone. This design enables the model to selectively refine spatiotemporal attention with a minimal parameter budget while the remainder of the network remains frozen.

To address visual inconsistencies at corruption boundaries, the framework employs two specialized refinement modules. First, a lightweight Boundary Refinement Head is introduced to operate on a morphological boundary band, which is computed via mask dilation and erosion. This head predicts an RGB residual and per-pixel blending weights to softly blend the restored content with the original uncorrupted pixels, effectively mitigating seam artifacts. Second, a lightweight U-Net-based Residual Refiner is utilized to predict a per-frame RGB residual on top of the initial model output. This refiner enforces strict data consistency by constraining its corrections exclusively to the masked corruption regions.

Finally, to maximize temporal consistency during inference, the pipeline incorporates a bidirectional Test-Time Augmentation (Reverse TTA) strategy. The video sequences are processed in both forward and reverse temporal directions, and the final predictions are obtained by merging the outputs using equal-weight averaging.

\begin{figure*}[ht]
\centering
\resizebox{\textwidth}{!}{%
\begin{tikzpicture}[
    node distance=1.0cm and 1.2cm,
    block/.style={rectangle, draw, fill=blue!8, text width=2.8cm, minimum height=1.0cm, align=center, rounded corners=3pt, font=\small},
    bigblock/.style={rectangle, draw, fill=orange!10, text width=3.2cm, minimum height=1.2cm, align=center, rounded corners=3pt, font=\small\bfseries},
    io/.style={rectangle, draw, fill=green!10, text width=2.5cm, minimum height=0.8cm, align=center, rounded corners=3pt, font=\small},
    arrow/.style={-{Stealth[length=2mm]}, thick},
    dasharrow/.style={-{Stealth[length=2mm]}, thick, dashed, color=gray}
]

% Input
\node[io] (input) {BSC Frames\\1280$\times$720};
\node[io, below=0.5cm of input] (mask) {Binary Masks\\1280$\times$720};

% Downsample
\node[block, right=1.5cm of input] (down) {Bicubic\\Downsample\\to 432$\times$240};

% B2SCVR
\node[bigblock, right=1.2cm of down] (b2scvr) {B2SCVR\\Restoration\\Model};

% B2SCVR sub-components (small)
\node[block, fill=purple!8, above=0.3cm of b2scvr, text width=2.0cm, minimum height=0.6cm, font=\scriptsize] (sam) {SAM2\\Encoder};
\node[block, fill=yellow!15, below=0.3cm of b2scvr, text width=2.0cm, minimum height=0.6cm, font=\scriptsize] (more) {MoRE\\Experts};

% Upsample
\node[block, right=1.2cm of b2scvr] (up) {Bicubic\\Upsample\\to 1280$\times$720};

% Compositing
\node[bigblock, fill=red!10, right=1.2cm of up] (comp) {Mask-Guided\\Compositing};

% Output
\node[io, right=1.2cm of comp] (output) {Restored\\Frames\\1280$\times$720};

% Arrows
\draw[arrow] (input) -- (down);
\draw[arrow] (down) -- (b2scvr);
\draw[arrow] (b2scvr) -- (up);
\draw[arrow] (up) -- (comp);
\draw[arrow] (comp) -- (output);

% Mask connections
\draw[dasharrow] (mask) -| (down);
\draw[dasharrow] (mask) -| (comp);

% Direct path for clean frames
\draw[arrow, color=green!60!black] (input) -- ++(0, -2.0) -| node[below, pos=0.3, font=\scriptsize, color=green!60!black] {Clean pixels (lossless copy)} (comp);

% SAM/MoRE connections
\draw[arrow, color=purple!60, thin] (sam) -- (b2scvr);
\draw[arrow, color=yellow!60!black, thin] (more) -- (b2scvr);

\end{tikzpicture}
}
\caption{Overview of weichow team pipeline. The B2SCVR model processes frames at 432$\times$240 (its native training resolution). Restored corrupted regions are upscaled to 1280$\times$720 and composited with original uncorrupted pixels via binary masks. Non-corrupted frames bypass the model entirely for zero quality loss.}
\label{fig:weichow}
\end{figure*}
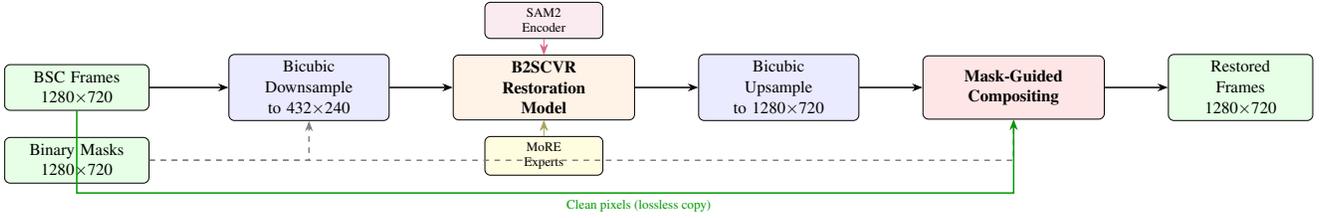

\textbf{Implementation details.}
The training and development process is structured into three distinct stages: individual module training, component integration with an exploration of temporal consistency losses, and final loss weight optimization. To ensure the robustness of the proposed method, the team curated a "worst10" validation subset consisting of the ten most challenging videos. This subset served as the primary evaluation metric, and only improvements that were empirically validated on these challenging cases were adopted.

During training, the model processes video clips consisting of 5 local frames and 3 reference frames, with patches cropped to a spatial resolution of $432 \times 240$. The network is optimized using the Adam optimizer coupled with Exponential Moving Average (EMA) using a decay rate of 0.999. The training is conducted with a batch size of 1 for a total of 30,000 iterations. A multi-step learning rate schedule is employed, with decay milestones set at 8,000 and 16,000 iterations and a decay factor ($\gamma$) of 0.5. The total loss function is formulated as a weighted sum of a hole reconstruction loss, a valid region reconstruction loss, and an $L_1$ regularization loss. To explicitly favor the restoration of corrupted areas, the corresponding loss weights are set to $\lambda_h=1.5$, $\lambda_v=0.5$, and $\lambda_{\ell_1}=0.1$, respectively.

For the testing phase, the framework employs a sliding-window inference strategy augmented with Reverse Test-Time Augmentation (TTA). Specifically, the sequence is processed in both forward and time-reversed directions, and the predictions are merged using equal-weight averaging. The boundary refinement module is applied independently inside each temporal pass, followed by a residual refiner that operates on the merged output. Finally, the restored frames are upsampled to their native resolution using bicubic interpolation, and a hard-mask compositing operation is applied to guarantee pixel-perfect fidelity in the uncorrupted regions.

\subsection{weichow}
\textbf{General method description.}
The weichow team proposes a video restoration approach that leverages the B2SCVR framework \cite{liu2025towards} as its core restoration backbone, as shown in Figure \ref{fig:weichow}. To effectively address the specific degradations of the challenge, the method introduces a mask-guided multi-resolution compositing pipeline. This specialized pipeline aims to preserve the original, undamaged content of the video without damage, while accurately reconstructing damaged areas through learning video repair. Additionally, the approach adapts the core architecture to the target corruption distribution by fine-tuning the base model on the challenge-specific dataset, thereby ensuring robust recovery of the degraded sequences.

\textbf{Implementation details.}
The proposed method leverages several pre-trained models, utilizing the B2SCVR model~\cite{liu2025towards} as the core restoration backbone, which was initially pre-trained on the BSCV dataset~\cite{liu2024bitstream}. To support the restoration process, the framework incorporates a frozen SAM2.1-tiny image encoder~\cite{ravi2024sam2} to provide multi-scale visual features, a frozen DINOv2-ViT-S/14 model~\cite{dinov2} for Mixture-of-Residual-Experts gating, and a frozen SPyNet~\cite{spynet} for bidirectional optical flow estimation. The total complexity of the model is 236.4M parameters, of which 56.8M are trainable, as the SAM2 encoder (179.6M parameters) and SPyNet (1.4M parameters) remain strictly frozen. No additional datasets are used for task-specific training beyond the official NTIRE challenge training set, which contains 3,471 video clips.

During the training phase, the pre-trained B2SCVR model is fine-tuned on the challenge data at a spatial resolution of $432 \times 240$. The network processes video clips consisting of 5 local frames and 3 reference frames. Optimization is performed using the Adam optimizer ($\beta_1=0$, $\beta_2=0.99$) with a batch size of 1. The learning rate is initialized at $2 \times 10^{-5}$ and follows a cosine annealing schedule down to $1 \times 10^{-6}$ over a total of 5,000 iterations. The overall loss function is a balanced combination of $L_1$ losses for both the corrupted and valid regions, formulated as $\mathcal{L} = 10 \cdot \mathcal{L}_1^{\text{hole}} + 10 \cdot \mathcal{L}_1^{\text{valid}}$. 

For testing, the inference pipeline operates in four distinct stages. First, the input frames and masks are read at their native $1280 \times 720$ resolution and downscaled to $432 \times 240$ for the model input, while the corruption masks are dilated using a $3 \times 3$ cross kernel for 4 iterations. Next, temporal video restoration is performed in chunks of 30 frames, utilizing a neighbor stride of 5 and a reference stride of 10. The backbone processes both clean and corrupted frame features, leveraging the SAM2 semantic priors and temporal propagation. In the third stage, the restored regions are upscaled back to $1280 \times 720$ via bicubic interpolation and composited using the binary mask, ensuring that the clean pixels from the original input are flawlessly preserved. Finally, a lossless clean frame handling strategy is applied: frames with a zero corruption mask are copied byte-for-byte from the input to prevent any quality degradation, while the restored corrupted frames are saved at a JPEG quality of 100. 

\begin{figure*}[t]
    \centering
    \includegraphics[width=\linewidth]{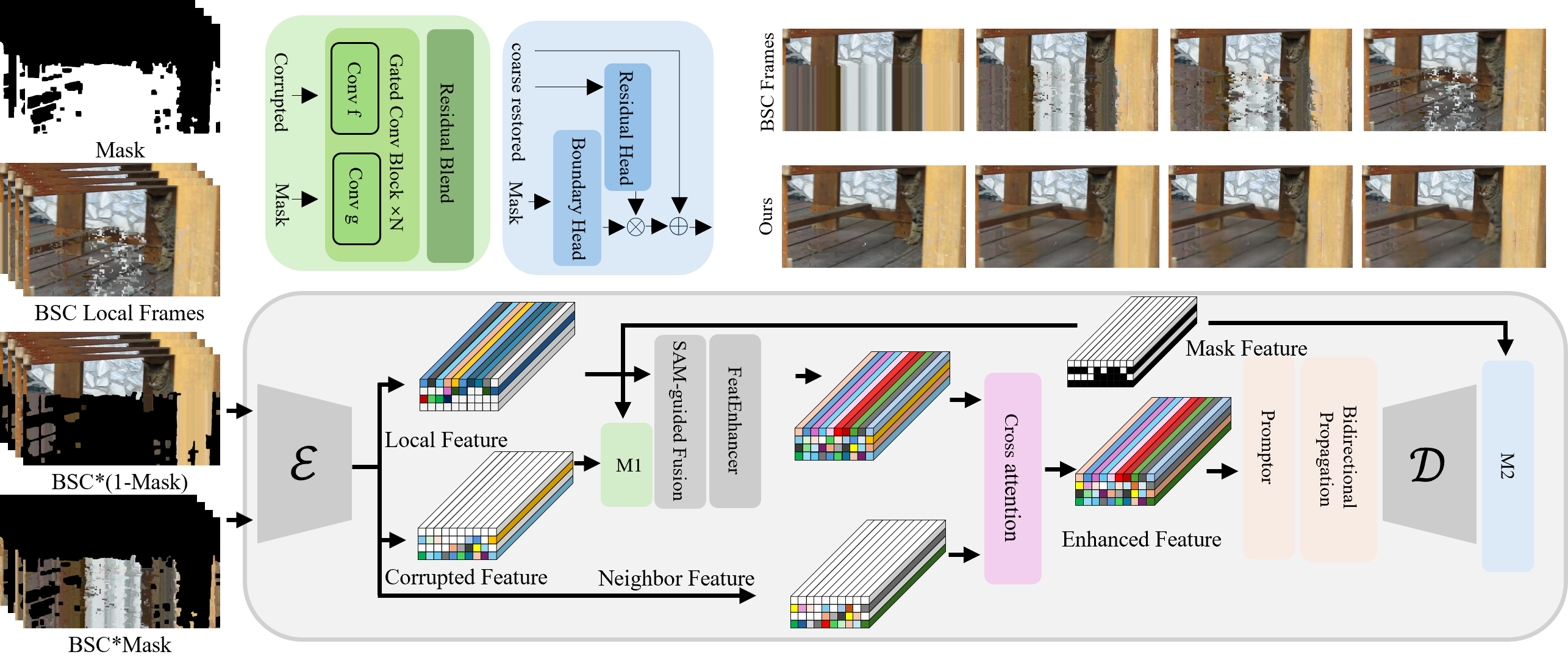}
    \vspace{-3mm}
    \caption{
    \textbf{The holding team: Beyond Missing Holes: Taming Feature Leakage in Mask-Guided Bitstream-Corrupted Video Recovery.}
    Built on a bidirectional propagation and temporal-transformer restoration backbone, our framework decomposes the input into valid and corrupted streams, suppresses unreliable corrupted-but-visible residuals with M1, retrieves cross-frame evidence through the target-centric cross-frame attention module (CA), and refines boundary seams with M2 after coarse restoration.
    }
    \label{fig:holding}
\end{figure*}

\subsection{holding}
\textbf{General method description.}
The holding team proposes a framework built upon the B2SCVR baseline~\cite{liu2025towards} for mask-guided bitstream-corrupted video restoration. The complete architecture and the interactions among the M1, CA, and M2 modules are illustrated in Figure \ref{fig:holding}. While existing pipelines utilize decoder-side masks to identify corrupted regions, they often suffer from feature leakage from these degraded areas, unstable temporal aggregation, and visible boundary artifacts. To address these limitations, the proposed method integrates three lightweight plug-in modules into the core architecture.

First, a Mask-aware Gated Suppression (M1) module is introduced to actively suppress corruption leakage during the initial feature extraction phase. Second, a Target-centric Cross-frame Attention (CA) mechanism is employed to enhance temporal aggregation by explicitly focusing the attention on the corrupted target regions across frames. Finally, a Boundary-aware Seam Refinement (M2) module is utilized to refine the visual transitions near the mask boundaries, effectively mitigating visible seam artifacts.

By incorporating these targeted modifications, the approach significantly improves both quantitative restoration performance and overall visual consistency while maintaining a practical, single-model inference pipeline. 

\textbf{Implementation details.}
The proposed framework is built upon the B2SCVR baseline~\cite{liu2025towards} and maintains a single-model architecture without relying on ensemble strategies, thereby ensuring a moderate computational complexity. The network is initialized with the official B2SCVR pre-trained weights. No additional external datasets are utilized; the model is fine-tuned exclusively on the official NTIRE 2026 BSCVR dataset using the provided video sequences and corresponding corruption masks.

During the training phase, the three newly introduced lightweight modules (for corruption suppression, temporal aggregation, and boundary refinement) are jointly optimized alongside the backbone network. The entire training process is conducted on a single NVIDIA H800 GPU and takes approximately 10 hours to complete.

For testing, the single trained model employs the provided decoder-side masks to guide the restoration of the corrupted regions. The inference process is efficient, requiring approximately 280 seconds to evaluate the entire test set on a single NVIDIA H800 GPU.

\subsection{NTR}
\textbf{General method description.}
The NTR team proposes a mask-guided video restoration approach built upon the B2SCVR framework. To effectively capture temporal context, the method processes a sliding window of consecutive local frames alongside evenly-spaced reference frames. Furthermore, to explicitly mitigate JPEG block-boundary artifacts, the provided corruption masks undergo a morphological dilation process prior to being utilized by the network.

The core generator architecture is designed as a multi-stage pipeline to ensure robust spatio-temporal recovery. First, a pre-trained SPyNet~\cite{spynet} module is utilized to estimate bidirectional optical flows. Next, a Convolutional Neural Network (CNN) encoder extracts base visual features, which are subsequently enhanced by fusing the masked and corrupted feature representations via a SwinIR-based module~\cite{swinir}. Following this feature extraction and fusion phase, the network performs bidirectional feature propagation using second-order deformable alignment, which is guided by the previously estimated optical flows. The propagated features are then passed through a series of Temporal Focal Transformer blocks~\cite{li2022towards} to perform comprehensive spatio-temporal aggregation. Finally, a CNN decoder reconstructs the restored frames, and the ultimate output is generated through a compositing operation, where the network's predictions are exclusively applied to the masked regions while the original uncorrupted pixels are preserved.

\textbf{Implementation details.}
The model is trained on a dataset comprising 3,471 videos from the BSCV benchmark~\cite{liu2024bitstream}, with frames resized to a spatial resolution of $432 \times 240$. To improve generalization, random horizontal flipping with a probability of 0.5 is applied as data augmentation. During training, each iteration samples a sequence consisting of 5 consecutive local frames alongside 3 reference frames.

The training process is divided into two distinct stages. In the first stage, the network is optimized for 250,000 iterations using the Adam optimizer ($\beta_1=0$, $\beta_2=0.99$) with a batch size of 3. The initial learning rate is set to $1 \times 10^{-4}$ and follows a MultiStepLR schedule, which decays the learning rate by a factor of 0.1 at the 200,000-iteration mark. The overall loss function for this generative phase combines $L_1$ losses for both the corrupted (hole) and uncorrupted (valid) regions (each with a weight of 10), an $L_1$ loss between the predicted and ground-truth optical flows, and a hinge GAN adversarial loss (with a weight of 0.01) computed by a spectrally-normalized 3D temporal patch discriminator.

The second stage focuses on PSNR refinement and spans an additional 60,000 iterations, resuming directly from the Stage 1 checkpoint. In this refinement phase, the adversarial discriminator is disabled to strictly optimize for distortion-oriented metrics. The learning rate is adjusted to $5 \times 10^{-5}$ and is decayed by a factor of 0.5 at 30,000 and 50,000 iterations using a MultiStepLR schedule, while maintaining the batch size of 3.During the testing phase, the framework employs a non-overlapping sliding window strategy, processing blocks of 5 local and 3 reference frames at the network's native $432 \times 240$ resolution. The restored outputs are subsequently upscaled to the target $1280 \times 720$ resolution utilizing Lanczos interpolation. In terms of computational efficiency, the inference pipeline requires approximately 22 seconds to process 25 test videos on a single GPU.
\section{Conclusion}
The NTIRE 2026 Bitstream-Corrupted Video Restoration (BSCVR) Challenge has successfully established standardized evaluation criteria for severe spatiotemporal artifacts and content distortion caused by transmission loss in the real world. In the team that submitted the final proposal, there was a development trend that emphasized both objective indicators and subjective perceptions. The MGTV-AI team has taken the lead in objective quantitative indicators such as PSNR and SSIM with the help of a three-stage progressive repair framework, demonstrating the advantages of multi-stage fusion in reconstructing structures; The RedMediaTech team made a breakthrough in visual perception quality and achieved the best LPIPS score. The team innovatively combined the Wan2.1 diffusion model with Qwen Image VAE, and successfully generated highly realistic textures and details in severely damaged segments through a two-stage training strategy, demonstrating the enormous potential of generative models in extreme repair tasks.

Looking at the technological trends of this competition, the combination of visual basic models and parameter efficient fine-tuning (PEFT) has become mainstream. Most teams widely introduce models such as SAM2 to extract advanced semantic priors, and utilize techniques such as LoRA to improve their generalization ability to complex damaged patterns while controlling computational costs. However, despite the powerful filling capabilities demonstrated by modern AI, the field still faces many challenges. When faced with extreme situations where native data is almost completely lost, models are still prone to bottlenecks such as edge softening, failure to reconstruct fine features (such as faces and text), and difficulty maintaining temporal coherence in large motion scenes. This challenge not only validates the potential of integrating spatiotemporal attention with generative models, but also points out a clear research direction for developing more robust and easily deployable video restoration systems in the future.

{
    \small
    \bibliographystyle{ieeenat_fullname}
    \bibliography{main}
}

\section*{Acknowledgements}
This work was conducted in the JC STEM Lab of Machine Learning and Computer Vision funded by The Hong Kong Jockey Club Charities Trust. This research received partially support from the Global STEM Professorship Scheme from the Hong Kong Special Administrative Region. This work was partially supported by the Humboldt Foundation. We thank the NTIRE 2026 sponsors: OPPO, Kuaishou, and the University of Wurzburg (Computer Vision Lab).
\appendix
\section{Teams and affiliations}
\subsection*{MGTV-AI}
\noindent\textit{\textbf{Title: }} A Three-Stage Framework For BitStream-corrupted Video Restoration \\
\noindent\textit{\textbf{Members:}} \\
Shiqi Zhou$^{1}$ (\href{mailto:shiqi@mgtv.com}{shiqi@mgtv.com}), \\
Xiaodi Shi$^1$ \\
\noindent\textit{\textbf{Affiliations: }} \\ 
$^1$ MGTV \\

\subsection*{RedMediaTech}
\noindent\textit{\textbf{Title: }} Bitstream-corrupted Video Restoration using One Step Wan2.1 \\
\noindent\textit{\textbf{Members:}} \\
Yuxiang Chen$^{1}$ (\href{mailto:chenyx.cs@gmail.com}{chenyx.cs@gmail.com}), \\
Yilian Zhong$^{1}$, \\
Shibo Yin$^{1}$, \\
Yushun Fang$^{1}$, \\
Xilei Zhu$^{1}$, \\
Yahui Wang$^{1}$, \\
Chen Lu$^1$ \\
\noindent\textit{\textbf{Affiliations: }} \\ 
$^1$ Xiaohongshu INC \\

\subsection*{bighit}
\noindent\textit{\textbf{Title: }} Two-Stage Bitstream-Corrupted Video Restoration via Semantic Memory and Mixture of LoRA Experts \\
\noindent\textit{\textbf{Members:}} \\
Zhitao Wang$^{1}$ (\href{mailto:zhitao.wang.hit@outlook.com}{zhitao.wang.hit@outlook.com}), \\
Lifa Ha$^{1}$, \\
Hengyu Man$^{1}$, \\
Xiaopeng Fan$^{1}$, \\
\noindent\textit{\textbf{Affiliations: }} \\ 
$^1$Harbin Institute of Technology \\

\subsection*{Vroom}
\noindent\textit{\textbf{Title: }} Enhanced B2SCVR: SAM2-Prior Guided Bitstream-Corrupted Video Restoration with LoRA and Boundary Refinement \\
\noindent\textit{\textbf{Members:}} \\
Priyansh Singh$^{1}$ (\href{mailto:2024uee0145@iitjammu.ac.in}{2024uee0145@iitjammu.ac.in}), \\
Krrish Dev$^{1}$,\\
Soham Kakkar$^{1}$,\\
Sidharth$^{1}$, \\
Dr. Vinit Jakhetiya$^{1}$, \\
Ovais Iqbal Shah$^{1}$\\
\noindent\textit{\textbf{Affiliations: }} \\ 
$^1$Indian Institute of Technology Jammu \\

\subsection*{weichow}
\noindent\textit{\textbf{Title: }} Mask-Guided Multi-Resolution Compositing with B2SCVR \\
\noindent\textit{\textbf{Members:}} \\
Wei Zhou$^{1}$ (\href{mailto:weichow@u.nus.edu}{weichow@u.nus.edu}), \\
Linfeng Li$^{1}$, \\
Qi Xu$^{2}$ \\
\noindent\textit{\textbf{Affiliations: }} \\ 
$^1$National University of Singapore \\
$^2$Shanghai Jiao Tong University \\

\subsection*{holding}
\noindent\textit{\textbf{Title: }} Beyond Missing Holes: Taming Feature Leakage in Mask-Guided Bitstream-Corrupted Video Recovery \\
\noindent\textit{\textbf{Members:}} \\
Zhenyang Liu$^{1}$ (\href{mailto:zyliu121426@163.com}{zyliu121426@163.com}), \\
Kepeng Xu$^{1}$ \\
Tong Qiao$^{1}$, \\
\noindent\textit{\textbf{Affiliations: }} \\ 
$^1$Xidian University\\

\subsection*{NTR}
\noindent\textit{\textbf{Title:}} Temporal Focal Transformer with Bidirectional Propagation for Bitstream-Corrupted Video Restoration \\
\noindent\textit{\textbf{Members:}} \\
Jiachen Tu$^{1}$ (\href{mailto:jtu9@illinois.edu}{jtu9@illinois.edu}), \\
Guoyi Xu$^{1}$ \\
Yaoxin Jiang$^{1}$, \\
Jiajia Liu$^{1}$, \\
Yaokun Shi$^{1}$, \\
\noindent\textit{\textbf{Affiliations: }} \\ 
$^1$University of Illinois Urbana-Champaign\\

\end{document}